\documentclass[runningheads]{llncs}
\usepackage{graphicx}
\usepackage{epsfig}
\usepackage{graphicx}
\usepackage{booktabs}
\usepackage{adjustbox}
\usepackage{multirow}
\usepackage{subcaption}
\usepackage{url}
\usepackage{tikz}
\usepackage{comment}
\usepackage{amsmath,amssymb} % define this before the line numbering.
\usepackage{color}

\usepackage[pagebackref=true,breaklinks=true,letterpaper=true,colorlinks,bookmarks=false]{hyperref}

\usepackage{xspace}
\makeatletter
\DeclareRobustCommand\onedot{\futurelet\@let@token\@onedot}
\def\@onedot{\ifx\@let@token.\else.\null\fi\xspace}

\makeatletter
\newcommand{\printfnsymbol}[1]{%
  \textsuperscript{\@fnsymbol{#1}}%
}

\def\etc{\emph{etc}\onedot} \def\vs{\emph{vs}\onedot}

\makeatother

\begin{document}
% \linenumbers
\pagestyle{headings}
\mainmatter
\def\ECCVSubNumber{26}  % Insert your submission number here

\title{\underline{ASAP}-NMS: \underline{A}ccelerating Non-Maximum Suppression Using \underline{S}patially \underline{A}ware \underline{P}riors} % Replace with your title

%\begin{comment}
\titlerunning{ECCV-20 submission ID \ECCVSubNumber} 
\authorrunning{ECCV-20 submission ID \ECCVSubNumber} 
\author{Anonymous ECCV submission}
\institute{Paper ID \ECCVSubNumber}
%\end{comment}
%******************

% % CAMERA READY SUBMISSION
% % \begin{comment}
\titlerunning{ASAP-NMS}
\author{Rohun Tripathi \thanks{Equal Contribution} \inst{1} \and
Vasu Singla \printfnsymbol{1} \inst{2} \and
Mahyar Najibi\inst{2} \and
Bharat Singh\inst{2} \and
Abhishek Sharma\inst{3} \and
Larry Davis\inst{2}}
\authorrunning{R. Tripathi et al.}
\institute{Cornell University, USA \and University of Maryland, USA \and Axogyan AI, India}

% \email{rt443@cornell.edu}
% % \end{comment}
% %******************
\maketitle

\begin{abstract}

The widely adopted sequential variant of Non Maximum Suppression (or Greedy-NMS) is a crucial module for object-detection pipelines. Unfortunately, for the region proposal stage of two/multi-stage detectors, NMS is turning out to be a latency bottleneck due to its sequential nature. In this article, we carefully profile Greedy-NMS iterations to find that a major chunk of computation is wasted in comparing proposals that are already far-away and have a small chance of suppressing each other. We address this issue by comparing only those proposals that are generated from nearby anchors. The translation-invariant property of the anchor lattice affords generation of a lookup table, which provides an efficient access to nearby proposals, during NMS. This leads to an \textbf{A}ccelerated NMS algorithm which leverages \textbf{S}patially \textbf{A}ware \textbf{P}riors, or \textbf{ASAP}-NMS, and improves the latency of the NMS step from 13.6ms to 1.2 ms on a CPU without sacrificing the accuracy of a state-of-the-art two-stage detector on COCO and VOC datasets. Importantly, ASAP-NMS is agnostic to image resolution and can be used as a simple drop-in module during inference. Using ASAP-NMS at run-time only, we obtain an mAP of 44.2\%@25Hz on the COCO dataset with a V100 GPU.
\end{abstract}

%%%%%%%%% BODY TEXT
\section{Introduction}
Highly-accurate real-time object detection pipelines are crucial for numerous practical applications such as autonomous driving, surveillance, robotics, medical image analysis and many more \cite{objectDetSurvey}.
These pipelines can be broadly classified into two categories - single-stage detectors (like SSD/YOLO/RetinaNet \etc \cite{liu2016ssd,redmon2017yolo9000,lin2018focal}) and two/multi-stage detectors (like Faster-RCNN/Mask-RCNN/ Cascade-RCNN \cite{ren2015faster,he2017mask,cai2017cascade} \etc). Two/Multi-stage detectors consistently outperform single-stage detectors in terms of accuracy and frequently appear as the winners of detection challenges \cite{peng2017megdet,liu2018path}. Owing to their superior accuracy, two-stage detectors are the workhorse for accuracy-critical industrial applications such as autonomous driving. Unfortunately, they are slower compared to single-stage detectors and this decreases their popularity for deployment to edge devices that require real-time processing. Therefore, improvements in the latency of two-stage detectors can lead to highly-accurate real-time object detection systems. 

%Specifically, Non-Maximum-Suprression, or NMS, is creating a latency bottleneck on the latest powerful GPUs \cite{Bolya2019YOLACTRI, Cai2019MaxpoolNMSGR}. \as{Why are we shying away from making it evident in the beginning that NMS is a reason behind slowing down proposal generation step? We have citations to show it and it catches the attention of reader by surprise.}

Two-stage detectors consist of a few common modules - the backbone network, region/object-proposal generation and bounding-box classification/ regression. The past few years have witnessed several innovations in backbone networks in the form of VGG, Inception, ResNet, MobileNet \cite{Simonyan2014VeryDC,2014GoingDW,he2016deep,Howard2017MobileNetsEC,shuffleNetV2} along with tremendous improvements in deep-learning specific compute power \cite{gpuCompute}. Together, they have resulted in significant improvement in object-detection pipelines - both in terms of accuracy and latency. However, the region-proposal generation module hasn't benefited much, \emph{in terms of latency}, from these advancements because it involves a greedy/sequential Non-Maximum-Suppression (NMS) step. Counter-intuitive as it may sound, but NMS for proposal generation is already turning out to be a latency-bottleneck for some recent object-detection pipelines \cite{Bolya2019YOLACTRI}; and will only get worse with faster hardware \cite{Cai2019MaxpoolNMSGR}. Profiling two-stage detectors \cite{ren2015faster,he2017mask} with different back-bones on different deep-learning libraries \cite{wu2019detectron2,Chen2019MMDetectionOM} with both GPU/CPU-NMS implementations support these observations and provide motivation for speeding-up NMS algorithm for proposal generation, Sec.\ref{sec:experiments}.

\begin{figure}[!t]
\centering
\includegraphics[width=1.0\linewidth]{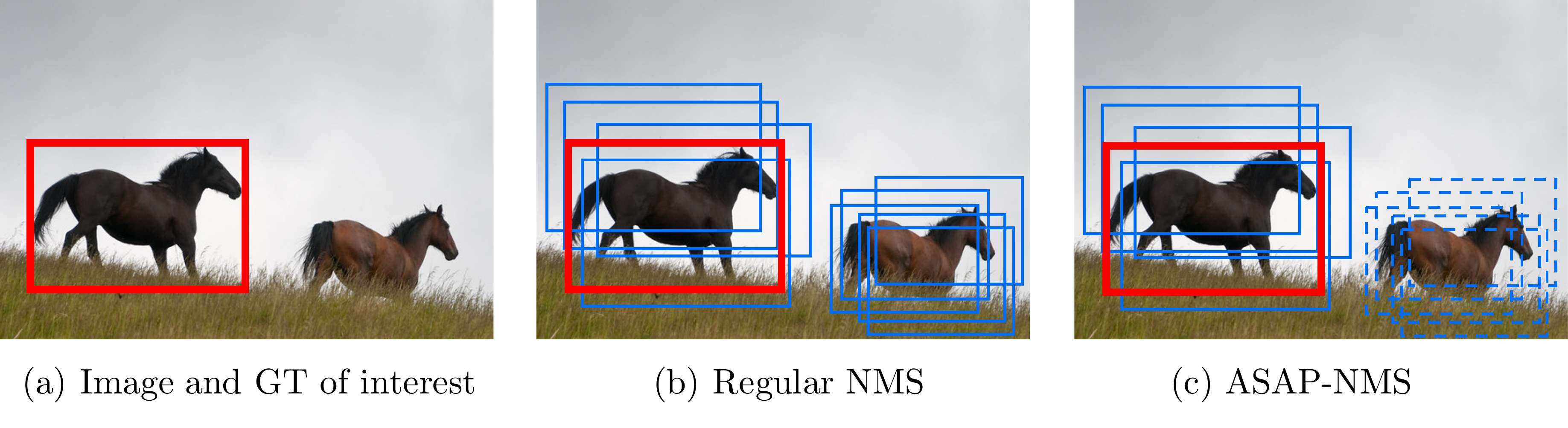}
\caption{(a) An image and a ground-truth object. (b) In regular NMS, the top-scoring box's (solid red) proximity is computed for \emph{all} other boxes (solid blue). (c) On the other hand, ASAP-NMS, exploits spatial priors to remove \emph{distant} boxes (dashed blue) from the proximity computation with the top-scoring box, which reduces the computational complexity of ASAP-NMS.}
\label{fig:teaser}
\end{figure}

The input to the NMS stage is a list of $\sim$10,000 top-scoring, pre-NMS, proposals based on their objectness score. The NMS algorithm iterates over the pre-NMS proposals, in decreasing order of objectness score, to compute the overlap (IoU) between the top-scoring proposal with the rest and removes the highly overlapping proposals, defined by an IoU threshold, from further consideration. This leads to an overall complexity of $\mathcal{O}(kN)$, where $N$ and $k$ are the pre-NMS and post-NMS numbers of proposals, respectively. Evidently, such a NMS algorithm ignores the spatial information available in the proposals and the top-scoring proposal is compared to every other proposal regardless of their relative locations in the image. Consequently, even far-away proposals are compared with each other at every iteration that leads to wasteful computations and calls for an optimization, see Fig.\ref{fig:teaser}.

While it's intuitively appealing to filter out far-away proposals from comparisons to reduce the complexity of NMS iterations, but it leads to a chicken and egg problem. Specifically, the cost of finding the far-away proposals, in terms of IoU, is the same as comparing all the proposals because the IoU still needs to be computed. This inspires the necessity of an efficient, pre-computable data-structure with $\mathcal{O}(1)$ lookup complexity to find the nearby proposals for a given proposal. Unfortunately, the image-dependent proposals, due to bounding-box regression and objectness score, cannot be pre-computed. Therefore, we need an \emph{efficiently pre-computable} source of \emph{spatial information} in the object-detection architecture that is \emph{invariant} to the image content. 

To this end,  we propose to leverage the spatial priors available from the anchor-grid/lattice to obtain a pre-computable lookup table of nearby proposals. Our choice of anchor-overlap as a proxy for proposal-overlap is motivated by that fact that the anchor-lattice is- (a) independent of the image-content and resolution, (b) affords efficient pre-computation of anchor-overlaps; the two properties needed to construct the required data-structure. This choice, however, raises further questions such as `would it lead to more false positives?' or `how would this affect the final mAP?'. To answer these, we empirically demonstrate that it doesn't lead to many false positives and note that such false positives will eventually be pruned away after the proposal-refinement stage. Therefore, it's unlikely to adversely affect the final detection mAP; a fact we demonstrate empirically. Intuitively as well, it makes sense because the bounding-box regression from the anchor-box to the proposal during RPN stage is a small refinement over the anchor and, therefore, far-away anchors will result in far-away proposals with a very high-probability. We name our proposed NMS algorithm as \textbf{A}ccelerated NMS using \textbf{S}patially \textbf{A}ware \textbf{P}riors, or \textbf{ASAP}-NMS for brevity.

The proposed ASAP-NMS is essentially an algorithmic improvement over Greedy-NMS that reduces the asymptotic computational complexity of NMS by reducing the number of required operations. Therefore, the speed-up doesn't depend on hardware acceleration or parallel-processing support which makes it favorable for a range of applications on edge devices that require fast and accurate detection.
% It requires a one-time pre-computation of a lookup table for nearby proposals that only depends on the anchor-placement scheme. 
% ASAP-NMS doesn't require re-training of the pipeline and affords simple inference-time integration with any two-stage object-detection pipeline, which makes it favorable for adoption in a wide range of applications on edge devices that require real-time detection capability. 
ASAP-NMS doesn't require re-training of the pipeline and affords simple inference-time integration with any two-stage object-detection pipeline, which enables wide adoption.
We empirically demonstrate 10x speed-up over Greedy-NMS for an open-source Multi-Scale Faster-RCNN object-detection pipeline that operates at high mAP of 44.2\% on the COCO dataset.

\section{Related Work}
NMS has served as a crucial component for numerous visual detection systems for the past 50 years, such as edge-detection \cite{rosenfeld1971edge}, key-point detection \cite{lowe2004distinctive,harris1988combined,mikolajczyk2004scale}, face-detection \cite{viola2001rapid}, and recently, object-detection \cite{2005HistogramsOO,felzenszwalb2010object,girshick2014rich}. In all the above cases, NMS is used to prune dense clusters of spatially-overlapping detection candidates. Such clusters are a common characteristic of any detection algorithm and they arise from the desired in-variance to small translations and deformations from the matching ``template''. Therefore, some form of application-specific NMS is necessary for any detection pipeline. Dalal and Triggs \cite{2005HistogramsOO} introduced the Greedy-NMS algorithm that suppresses nearby detection candidates w.r.t. a high-scoring candidate based on an overlap threshold. Since then, Greedy-NMS has been the de-facto standard NMS algorithm for object-detection pipelines \cite{girshick2015fast,ren2015faster,liu2016ssd,redmon2016you,li2017light}. This version has been adopted in almost all popular object-detection pipelines both in academia and industry owing to its simplicity and modular nature.

%As discussed in the previous section, convolutional object detection systems are either single-stage, or proposal-free architectures, or the proposal-based two-stage architectures. YOLO \cite{yolo}, SSD \cite{ssd}, RetinaNet \cite{retina} are some representative examples of single-stage detection systems that produce classification and bounding-box regression on a densely sampled grid of ROIs sampled throughout the image at different scale and aspect-ratio. \as{is NMS as crucial for single-stage detectors as it is for two-stage detectors too?} On the other hand, the proposal-based two-stage detectors first propose a set of densely packed candidate set of proposals, $\mathcal{S}$, by classifying and regressing anchor-boxes at different scales and aspect-ratio. This candidate set further undergoes Non-Maximal Suppresion to select the set of proposals, $\mathcal{P}$ which are fed to the second stage for further refinement. Different variants of RCNN architecture \cite{rcnn} such as, Fast-RCNN \cite{fast-rcnn}, Faster-RCNN \cite{faster-rcnn}, Cascade R-CNN \cite{cascade-rcnn} are some representative systems in this category. R-FCN \cite{rfcn} is another popular architecture that aims at reducing the computation during refinement by sharing the compuatation via convolution layers for classification and regression over a dense-grid instead of the proposal set $\mathcal{P}$. Typically, two-stage architectures outperform the single-stage architectures in terms of accuracy but they do worse in terms of computation time.
One would expect that since NMS is such a crucial module for object-detection, it would have received a lot of attention to improve its accuracy and latency. In fact, it has largely been ignored in favour of efforts to improve other modules in object-detection pipelines and only in the past couple of years it has gathered some traction. Recently, Soft-NMS \cite{bodla2017soft} analyzed the behaviour of NMS for strongly overlapping objects and proposed to only decay the score of proposals instead of completely removing them, which results in improvements of 1-2\% in mAP across different datasets. %It doesn't require any re-training and can be used as a drop-in replacement for Greedy-NMS at inference-time; it has been adopted in most standard detection libraries. 
Other recent approaches towards improving the performance of NMS are - Fitness NMS \cite{TychsenSmith2017ImprovingOL} which uses an IoU-weighted classification score, TNet \cite{HosangBS15Tnet} which employs a convnet for NMS, GossipNet \cite{HosangBS17GNet} which exploits pair-wise contextual features between proposals, Relation Net \cite{HuHanRelationNet} which computes relation features for detections based on the image appearance, IoU-Net \cite{JiangG18IOUNet} which learns IoU from the ground-truth and uses it for NMS, Uncertainty-NMS \cite{uncertNMS} which exploits the uncertainty associated with bounding-boxes to merge nearby boxes during NMS, and Learning NMS \cite{hosang2017learning} which uses features computed on detection boxes and their scores for NMS. All these approaches have resulted accuracy improvements over Greedy-NMS, however, their focus hasn't been on improving the speed of NMS, which is the focus of this work. Due to computational efficiency reasons, the aforementioned methods are favorable to the second-stage NMS mostly, on 300-1000 proposals to produce the final detection boxes. %Also, they cannot be used as a drop-in module during inference time only.  %Therefore, they have witnessed significantly lesser adoption in open-source libraries vs. Soft-NMS. A strong indication of desirability for inference-time only modules. - Lets uncomment this for camera ready, some reviewer my might get pissed 

%however, their focus hasn't been on improving the speed of NMS.

%Among other factors such as more computation, NMS can be a potential bottleneck to speed of two-stage architectures as well, as pointed out by \cite{Cai2019MaxpoolNMSGR}. Greedy-NMS has been a essential component of detection systems for more than a decade since it was first proposed in \cite{dalaltriggs}. For quite a long time, the greedy-NMS has been the de-facto algorithm for selecting the proposals and final detections from a candidate set. Up until recently, greedy-NMS was the only implementation of NMS in almost all the popular object-detection libraries such as, Detectron, TensorFlow Detector, OpenCV, \as{please add more}. A significant departure from greedy-NMS came in the form of Soft-NMS \cite{soft-nms}, which carefully analyzed the behaviour of NMS under strong clutter and overlapping objects and proposed to only decay the score of proposals instead of completely removing them. It consistently improved the performance of a majority of object detection systems on popular benchmarks by 1-2\% mAP. Today, Soft-NMS has been adopted to almost all the standard detection libraries due to its completely modular and drop-in nature at the inference time. Inspired by the success of Soft-NMS, a few other notable approaches such as, Fitness NMS \cite{TychsenSmith2017ImprovingOL}, TNet \cite{tnet}, GNet \cite{gnet}, Relation Net \cite{relationNet}.

With more computationally efficient backbones and evermore powerful GPUs, NMS has became a significant latency overhead, as shown in \cite{Cai2019MaxpoolNMSGR,Bolya2019YOLACTRI}. Therefore, some efforts have also been made to reduce the latency of NMS. MaxPool-NMS \cite{Cai2019MaxpoolNMSGR} exploits spatial max-pooling among nearby proposals for NMS which results in impressive speed-up at the cost of negligible degradation in performance. YOLOACT \cite{Bolya2019YOLACTRI} allows already discarded boxes to suppress other boxes which results in a parallel Fast-NMS approach that is convenient for GPU operations (assign, threshold, max) on an $N \times N$ matrix. It results in a significant speed-up with marginal performance degradation. Unfortunately, the speedup achieved by YOLOACT is still not sufficient to catch up with the latency reduction in neural networks due to better architectures and faster GPUs. Unlike these approaches, which rely on GPU-compute for speed-up or lead to performance degradation, we exploit the existing spatial structural priors in the anchor-space to achieve similar speed-ups for NMS with no accuracy drop. ASAP-NMS inherently reduces the total number of comparisons in NMS to reduce the complexity of the operation. It is a fundamentally different way of thinking how NMS can be performed when priors about the anchor lattice are available.

%Therefore, our approach is also applicable to single-process compute platforms.

\section{Background}
\begin{figure}[!t]
\centering
\includegraphics[width=\linewidth]{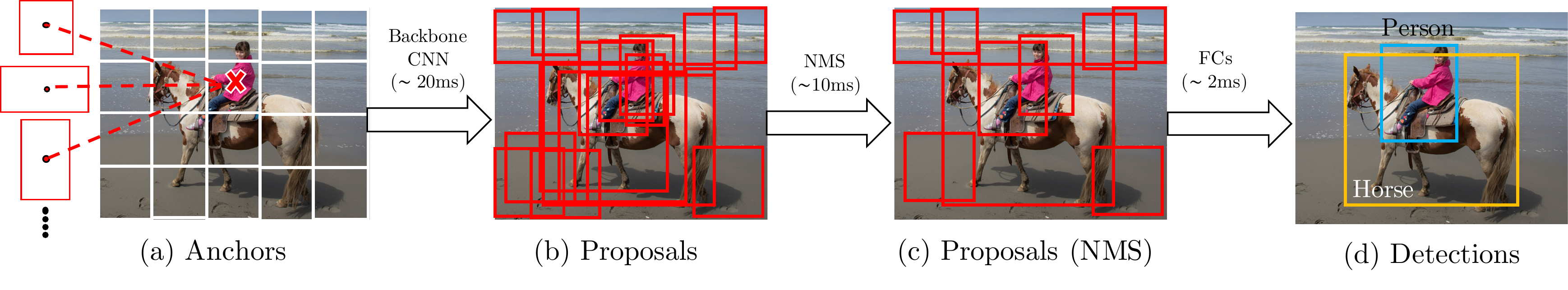}
\caption{Object detection pipeline and processing times in ms for an 800x1280 image for 300 post-NMS proposals. (a) A dense regular grid of $\sim$50,000 proposals anchors placed over the image. (b) From anchors to $\sim$50,000 proposals after regression. (c) NMS on the top-scoring $\sim$10,000 proposals to remove redundant proposals. (d) RoI-Pooling for the remaining 300 proposals followed by application of fully connected layers and final NMS to obtain the final detections.}
\label{fig:pipeline}
\end{figure}

When the two-stage end-to-end architecture for object detection was first introduced in 2015 \cite{ren2015faster}, it used high-dimension features for proposals along with a computationally heavy refinement-head. For example, using VGG-16 \cite{Simonyan2014VeryDC} as the backbone architecture resulted in a 512-dimensional conv5 feature map, which after RoIPooling outputs a $7\times7\times512$ blob. In order to facilitate the use of pre-trained weights during fine-tuning, the same $4096$-dimensional FC layer from VGG-16 was used in the refinement-head of Faster-RCNN. Therefore, when the number of proposals was increased to 500 or 1000, the computational overhead of the hidden-layer that transforms the $7\times7\times512$ input blob to $4096$-dimensional feature increased significantly. Fortunately, as the back-bone architectures improved over time, it was observed that even the computationally-lighter Faster-RCNN architectures obtained similar performance. For example, we can compress the conv5 features of ResNet-101 from $2048$ to $256$ before performing RoIPooling. Also, the size of the FC layer can be changed to only $1024$. Together, these changes reduce the FLOps of the Faster-RCNN refinement-head by 8 times. On modern GPUs, the Faster-RCNN head is no longer a computational bottleneck, which can be seen from the timings presented for the CNN backbone and the FC layers for different Faster-RCNN backbones in Fig \ref{fig:pipeline}.

%Whats the problem now?
However, as GPUs have become faster over time, the intermediate step of processing the scores assigned to the anchor boxes to generate a limited number of proposals has become a computational bottleneck in two-stage object detectors, Fig \ref{fig:pipeline}. For 300 proposals, the proposal generation step can take as long as $6-12 ms$ (depending on the detection library), which is significant when compared to the inference-time of ResNet-50 on a $800 \times 1200$ resolution image ($\sim 20 ms$), on a V100 GPU! Why so? As it turns out, the NMS algorithm used in the proposal generation step is responsible for the slow-down. Further details of proposal generation are discussed in the next section.

\section{ASAP-NMS}
This section first explains region-proposal network, abbreviated as RPN, and the importance of NMS for proposal generation. Then, it highlights the shortcoming of Greedy-NMS in terms of latency followed by a detailed description of the proposed spatial prior, its incorporation into ASAP-NMS and a strided version of ASAP-NMS for bounded performance guarantees.

\subsection{RPN and Greedy-NMS}
%describe the existing NMS used for proposal generation, reference an algorithm chart - use similar stuff from the intro
The Region Proposal Network (RPN) assigns an objectness score to each anchor-box which is placed on the image. For an image of size $W \times H$ pixels, a total of $\frac{W}{S} \times \frac{H}{S}$ anchor-placement locations are obtained when the stride of the CNN is $S$. Given an anchor-template set $\mathcal{T} = \{T^1, T^2, ... T^k...T^K\}$, corresponding to different scales and aspect-ratios, a total of $K \times \frac{W}{S} \times \frac{H}{S}$ anchor-boxes, $\mathcal{A}$, are placed over the image. Every element in $\mathcal{A}$ is assigned an objectness score by the RPN. Therefore, for a $1024\times1024$ pixels image, a CNN of stride 16 and $15$ anchors per location, generates a total of $61,440$ proposals. Refining this many proposal-boxes even with a lightweight Faster-RCNN head is computationally demanding. For example - the lightweight head mentioned in the previous section would take $\sim$800 GFLOps to refine $61,440$ proposals vs. $\sim$200 GFLOps for the entire ResNet-101 backbone! Generally, a significant portion of natural images corresponds to the background; therefore, a large fraction of anchor-boxes can be easily filtered out by a simple score-threshold. Typically, the top-scoring set of $\sim$10,000 bounding-boxes, $\mathcal{S}$, is retained for post-processing as the remaining anchor boxes are not likely to contain objects. Finally, NMS is applied on $\mathcal{S}$ to remove spatially redundant proposals that are very close to each others while ensuring high recall for all the objects in the image with a limited candidate set.

\begin{figure}[!th]
\includegraphics[width=0.8\linewidth]{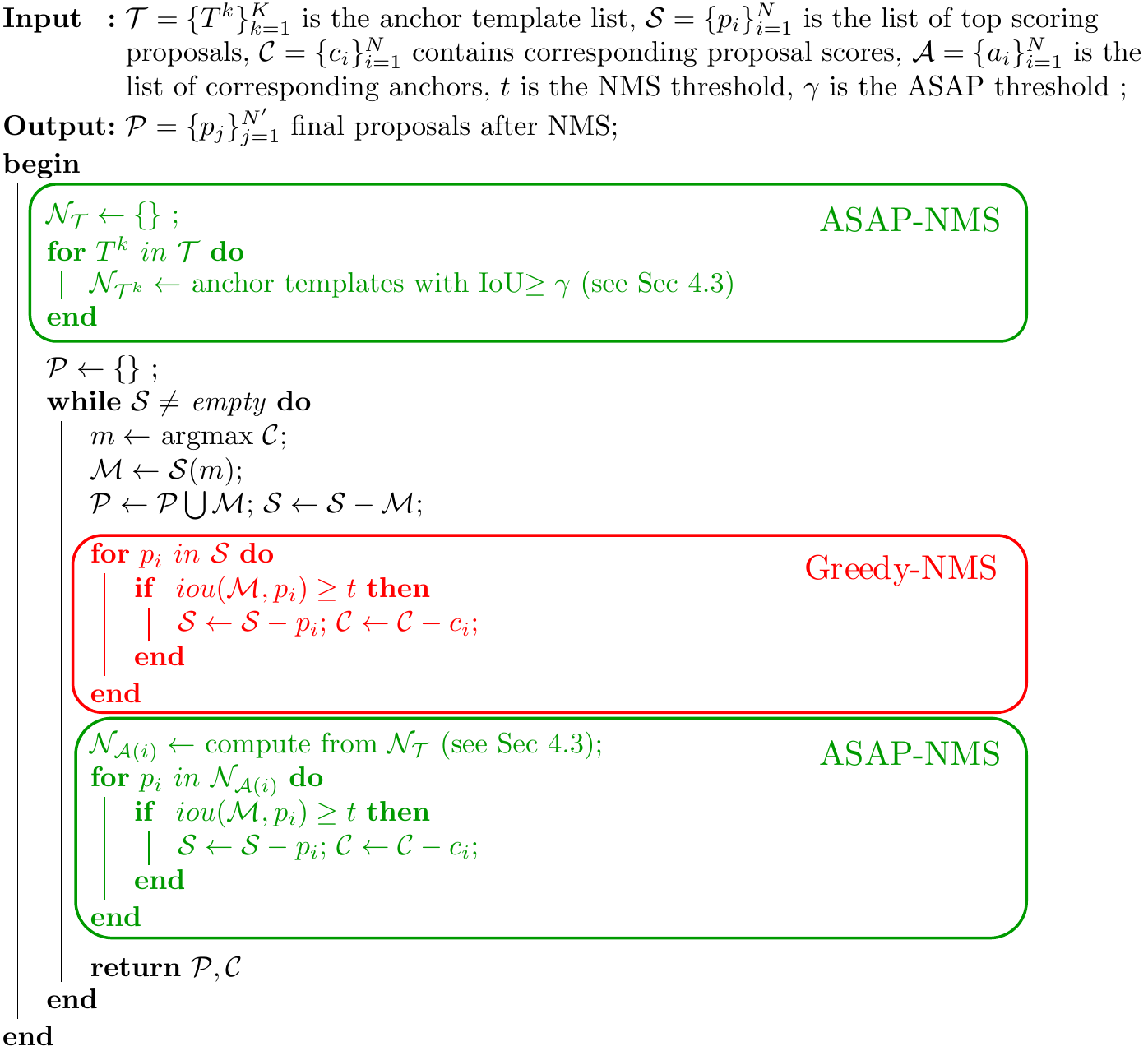}
\caption{Pseudocode for ASAP-NMS compared with greedy NMS. We replace the red block in greedy NMS with green blocks. ASAP-NMS speeds-up the process by reducing the number of IoU comparisons.}
\label{fig:asap_algo}
\end{figure}

The popular Greedy-NMS algorithm is sequential in nature and computationally expensive. At each iteration $i$, it selects the top scoring proposal $P(i)$ from the set $\mathcal{S}$ and removes all proposals in $\mathcal{S} - P(i)$ which have an overlap $o$ greater than a threshold $t$. Hence, the complexity of each iteration is linear in the size of set $\mathcal{S}$, see Fig \ref{fig:asap_algo}. The filtering step which reduces the size of the set from $K \times \frac{W}{S} \times \frac{H}{S}$ to $\mathcal{S}$ is crucial to achieve an acceptable run-time. Also, the complexity of Greedy-NMS increases linearly with the number of selected proposals. Unfortunately, the $i+1^{th}$ iteration depends on the output of the $i^{th}$ iteration; hence, parallel computation of GPUs cannot be leveraged trivially to accelerate it. The total complexity turns out to be $\mathcal{O}(|\mathcal{S}|\times|\mathcal{P}|)$, where $\mathcal{P}$ is the post-NMS set of proposals. The typical cardinality of $\mathcal{S}$ and $\mathcal{P}$ used in popular state-of-the-art object detection pipelines are $\sim$10K and $\sim$300-1000, respectively.

\subsection{Spatial Priors for ASAP-NMS}
%explain why spatial priors are important for NMS, explain Fast-NMS visually with a figure and an algorithm
From above, we note that $|\mathcal{S}| \gg |\mathcal{P}|$. Moreover, a significant reduction in the size of the post-NMS proposal set $\mathcal{P}$ can adversely affect the final object detection accuracy (especially for cases like face detection which can have up to a 1000 faces in an image). Hence, the only possibility to reduce computation without sacrificing accuracy is by reducing the complexity of each iteration. We questioned whether it is necessary to compare the top-scoring proposal to all the proposals in $\mathcal{S}$ and will explain, in fact, why it is not. For example, in Fig \ref{fig:teaser} two distant proposals containing two different objects are still compared by Greedy-NMS. Obviously, such comparisons are superfluous. Moreover, NMS for RPN module in two-stage detectors is applied with a large overlap threshold $t=0.7$ - aimed at removing strongly overlapping proposals. Therefore, we can safely ignore a comparison between two proposals, $P_m$ and $P_n$, whose corresponding anchors, $A_m$ and $A_n$, have a low overlap in the anchor-space. To check the validity of this claim, we report the probability of any pair of proposals' overlap exceeding $0.7$ as a function of the overlap of their associated anchor-boxes, in Table \ref{tab:flipping}. Note that even for an anchor overlap ranging between 0.2 and 0.3, the probability of the corresponding proposal overlap exceeding 0.7 is only 0.025. The existence of anchor pairs that have an anchor overlap $\le$ 0.3 but a proposal overlap $\ge$ 0.7 can potentially result in retaining both the proposals after NMS, because such proposals will not be compared during NMS. It can potentially lead to spatially redundant proposals after NMS.

\begin{figure}[!t]
\centering
\includegraphics[width=\linewidth]{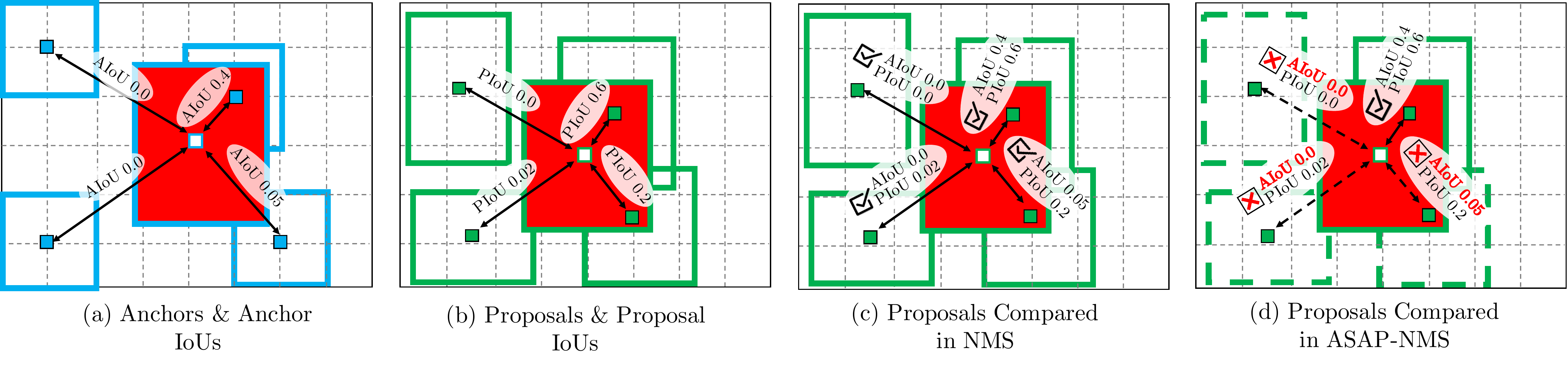}
\caption{ASAP-NMS \vs greedy NMS. (a) Anchors on a regular grid and their IoU. The anchor corresponding to the high scoring proposal is shown in red and the rest are shown in blue. (b) The proposals (green boxes) generated by moving and scaling the corresponding blue anchors. The high scoring proposal is shown in red. (c) NMS compares the red high-scoring proposal (red box) with all others (green boxes), regardless of the initial IoU between their corresponding anchors. (d) ASAP-NMS only compares the high scoring red box with those that have a reasonable overlap in the anchor space.}
\label{fig:fastnms_vs_nms}
\end{figure}

\begin{table}[!ht]
\centering
\setlength\tabcolsep{8pt} 
\begin{tabular}{|c|c|c|c|c|c|}
\multicolumn{6}{c}{Initial IoU Range}\\
\hline
 0.0-0.1 & 0.1-0.2 & 0.2-0.3 & 0.3-0.4 & 0.4-0.5 & 0.5-0.6  \\
\midrule
0.03\% & 0.8\% & 2.5\% & 7.6\% & 15.5\% & 26.4\% \\
\bottomrule
\end{tabular}
\caption{Percentage of anchors on COCO with an initial anchor IoU in a given range having a proposal IoU of more that 0.7 after bounding box regression.}   
\label{tab:flipping}
\vspace{-5mm}
\end{table}

However, even if such proposals are not suppressed at this stage, they will eventually get suppressed during NMS at the refinement-stage. Therefore, it is not necessary to achieve perfect NMS results in the first stage and the cost of having a small percentage of false positives is not high. To this end, at the $i^{th}$ iteration, we only compare the proposal $P(i)$, originating from the anchor $A(i)$, with a set of proposals, $\mathcal{N}_{A(i)}$, whose anchors have a significant anchor overlap with $A(i)$, while safely ignoring remaining proposals, see Fig \ref{fig:fastnms_vs_nms} and \ref{fig:asap_algo}. This idea hinges upon the premise that only highly overlapping anchors are likely to result in highly overlapping proposals that need to be suppressed for removing spatial redundancy during NMS. Therefore for each iteration, $|\mathcal{N}_{A(i)}| \ll |\mathcal{S}|$, which results in a significant reduction in computation. It gives rise to a generalized NMS algorithm that restricts the overlap comparison among proposals within spatially nearby neighbourhoods to reduce computation. The next section introduces a scheme to efficiently obtain the neighbourhoods for each proposal with the help of lookup tables.

%The premise is that only nearby anchors are likely to get suppressed during proposal generation after bounding box regression and $|\mathcal{N}_A(i)|$ is much smaller than $|\mathcal{S}|$. We show this is indeed true empirically in our experiments. %that the size of $\mathcal{N}_a(i)$ for different anchor overlap thresholds for two different anchor placement schemes in Table \ref{} \textcolor{red}{put sizes for 9 and 15 with 0.1 delta}. 

\subsection{Spatially Invariant Anchor Templates}
%approach for resolution agnostic anchor templates
Here, we show how to obtain $\mathcal{N}_{A(i)}$ on-the-fly with the use of a pre-computed anchor neighborhood table $\mathcal{N}_{\mathcal{T}}$. Due to the spatially uniform lattice-like structure of the anchor grid, the neighbourhood table $\mathcal{N}_{\mathcal{T}}$ is invariant to the resolution of the image and the spatial location of the anchors in the image. Hence, $\mathcal{N}_{\mathcal{T}}$ can be pre-computed in the form of a lookup table. Therefore, obtaining $\mathcal{N}_{A(i)}$ from $\mathcal{N}_{\mathcal{T}}$ just involves a simple lookup operation from the proposal to its corresponding anchor index. However, we still need to check whether the anchors corresponding to the proposals in $\mathcal{N}_{A(i)}$ fall inside the image or not.

% We now describe the method to construct $\mathcal{T}_a(i)$ more formally. Let $a_k$ denote an 
We now describe the details to construct $\mathcal{N}_{A(i)}$ from $\mathcal{N}_{\mathcal{T}}$. First, let's take a look at the IoU formulation between two anchors, $T^k$ and $T^l$, as a function of their height and width, $[(h^k, w^k), (h^l, w^l)]$, and the displacement between their centers, $\delta_x$ and $\delta_y$-
\begin{gather*}
IoU(T^k, T^l) = \frac{Area(T^k \cap T^l)}{Area(T^k) + Area(T^l) - Area(T^k \cap T^l)} \\
(T^k \cap T^l) = max(x_{2}^i - x_{1}^i, 0) * max(y_{2}^i - y_{1}^i, 0) 
\end{gather*}
\begin{align*}
x_{1}^i &= max(-w^k/2, \delta_x - w^l/2) \hspace{8mm}
x_{2}^i = min(w^k/2, \delta_x + w^l/2) \\
y_{1}^i &= max(-h^k/2, \delta_y - h^l/2) \hspace{8mm}
y_{2}^i = min(h^k/2, \delta_y + h^l/2) \\
\end{align*}

The formulation only depends on the type of anchors and the relative distance between them that makes it \emph{translation-invariant}. In other words, the IoU between any two anchors doesn't depend on their absolute location. Since, the anchors are placed uniformly on a spatial lattice, the distances between them form a discrete set of elements. Note that the IoU between any two anchors decreases with increasing relative displacement between them. Therefore, the relative displacement from a given anchor, $T^k$, for all the anchors with IoU $\ge \gamma$ is bounded by some $(\delta_x^{\gamma}$, $\delta_y^{\gamma})$, which can be computed for any given value of $\gamma$. Next, we define $\mathcal{N}^{\gamma}_{A}$ as the set of anchors $\{A_j : IoU(A, A_j) \ge \gamma\}$ or equivalently, $\{A_j : Rel(A, A_j) \le (\delta_x^{\gamma}, \delta_y^{\gamma})\}$, where, $Rel(A, B)$ is the relative displacement between anchor $A$ and $B$. The fact that both the anchors-types, and relative displacements are finite sets of elements, we have $|\mathcal{N}^{\gamma}_{A}| \sim \mathcal{O}(K \times \delta_x^{\gamma}/s_{\mathcal{A}} \times \delta_x^{\gamma}/s_{\mathcal{A}})$, where $s_{\mathcal{A}}$ is the size of the anchor stride. Hence, for a given $A$, $\gamma$ and anchor-template set $\mathcal{T}$, a pre-computable finite set of displacements will yield all the elements in $\mathcal{N}^{\gamma}_{A}$. This aforementioned set of displacements is stored in the form of an efficient lookup table to obtain $\mathcal{N}^{\gamma}_{A}$. Now, exploiting the translation-invariant property of $\mathcal{N}^{\gamma}_{A}$ and fixing $\gamma$, we obtain the required anchor neighbourhood table $\mathcal{N}_{\mathcal{T}}$. Finally, we can obtain $\mathcal{N}_{A(i)}$ from $\mathcal{N}_{\mathcal{T}}$ by a single lookup operation to go from proposal $P(i)$ to the corresponding anchor $A(i)$, followed by using the stored lookup table $\mathcal{N}_{\mathcal{T}}$ to fetch the list of neighbourhood anchors and their proposals.

% We now describe the method to construct $\mathcal{T}_a(i)$ more formally. For any  $a_{k}$, $\mathcal{T}_{a_k}$ consists of all anchors which have an overlap greater than $\gamma$ with all anchor sets $A_{\delta_x,\delta_y}$ placed at a distance $\delta_x$ and $\delta_y$ away from the anchor $a_k$.  \textcolor{red}{explain in more detail, why spatially invariant etc...}

\begin{table}[!b]
\centering
\scriptsize
\setlength\tabcolsep{6pt} 
\begin{tabular}{c|c|c|c||c|c||c|c||c|c||c|c}
\multirow{2}{*}{$\gamma$} & \multirow{2}{*}{2} & \multirow{2}{*}{4} & \multirow{2}{*}{7} & \multicolumn{2}{c||}{13} & \multicolumn{2}{c||}{24} & \multicolumn{2}{c||}{time} & \multicolumn{2}{c}{mAP} \\ \cline{5-12} 
 &  &  &  & R & S & R & S & R & S & COCO & VOC \\ \toprule
Greedy   & 12k & 12k & 12k & 12k & - & 12k & - & 13.6 & - & 44.3 & 86.0 \\
0.1 & 75 & 387 & 1k & 3.5k & 2.1k & 5.6k & 2.8k & 9.7 & 5.2 & 44.3 & 86.0\\ 
0.2 & 31 & 152 & 487 & 1.6k & 1k & 2.9k & 1.3k & 5.7 & 3.1 & 44.2 & 86.0\\ 
0.3 & 7 & 56 & 148 & 445 & 393 & 1.3k & 382 & 2.7 & 1.7 & 44.2 & 86.0\\ 
0.4 & 3 & 17 & 45 & 159 & 159 & 561 & 136 & 1.7 & 1.2 & 44.2 & 85.9\\ 
0.5 & 1 & 6 & 25 & 88 & 88 & 295 & 75 & 1.4 & 1.1 & 44.1 & 85.9\\
0.6 & 0 & 3 & 5 & 24 & 24 & 81 & 21 & 1.1 & 1.0 & 43.9 & 85.9\\
0.7 & 0 & 1 & 3 & 12 & 12 & 40 & 9 & 1.0 & 0.9 & 43.7 & 85.9\\
0.8 & 0 & 0 & 1 & 4 & 4 & 14 & 4 & 0.9 & 0.9 & 43.0 & 85.7\\
\end{tabular}
\caption{Size of $\mathcal{N}_{\mathcal{T}^k}$ for anchors of different anchor scales and NMS run-time on CPU (ms, column $time$) for a 800x1280 image for different values of $\gamma$. Columns 2, 4, 7, 13, 24 represent the scales of the area of the anchors at a stride of 16. Column (R) represents dense anchor placement and (S) represents strided anchor placement.
%For $\gamma >=0.3$, even for very large anchors, $|\mathcal{T}|$, is considerably smaller than GreedyNMS.
mAP values for MS-COCO and VOC are reported for inference on 2 scales. Performance is the same for both anchor placement strategies.}
\label{tab:AnchorList}
\end{table}
\subsection{Adaptive stride for Anchor Placement in ASAP-NMS}
%Describe the variable stride for different anchors in Fast-NMS, use a Figure
$\mathcal{N}_{A(i)}$ is essentially a list of anchors that are close to the anchor $A(i)$. Therefore, this list will be large if more anchors are placed close to each other on the anchor placement grid. In order to reduce the number of nearby anchors, we increase the anchor-stride for large scale anchors, $384\times384$ pixels, which typically get associated with large objects.
% \rt{ We don't mention how much the stride is increased to. This is mentioned in 5.6}
Typically, the size of $\mathcal{N}_{A(i)}$ for large anchors is also large due to their size, see Table \ref{tab:AnchorList}. Since, most of the anchors placed close to the center of a large object are good candidates for accurate detection, we don't require multiple large anchors close to each other. Therefore, this modification does not affect the accuracy of the algorithm but reduces the size of $\mathcal{N}_{A(i)}$ for large anchor boxes.

\section{Experiments} \label{sec:experiments}
%Tear fast-nms down and explain why it works.
% \rt{do we need a summary of the experiments section before setup?}
% Do we have the space for it?

%\subsection{Experimental Setup}
 For our experiments we use different values of $\gamma$ and show the effect of $\gamma$ on $|\mathcal{N}_{\mathcal{T}}|$ and latency. We use two different Faster-RCNN open-source detection libraries: Multi-Scale Faster-RCNN (SNIPER) \cite{sniper2018} and Detectron2 \cite{wu2019detectron2}. We used 2 scales for SNIPER at inference time which leads to an mAP of 44.3\% on the COCO val2017 set. With Detectron2, we perform single scale inference which leads to an mAP $\sim$ 40\%. We use ResNet-50 and ResNet-101 as the backbones and the default training settings in the respective libraries. During inference, we generate anchors at 5 scales and 3 anchor ratios leading to a total of 15 anchors per anchor position. Following \cite{sniper2018}, we only apply NMS to the top 12,000 top scoring proposals and use an NMS threshold of $0.7$ for proposal generation. Ablation experiments are performed with \cite{sniper2018}.

% GreedyNMS selects top-k highest scoring proposals (typically 12000) and suppresses proposals which have a high IOU ($>0.7$) with current top scoring proposal against every other proposal.
% Using ASAP-NMS, we are able to reduce number of comparisons for every proposal from k to $|\mathcal{T}_{a^k}|$. The size of $|\mathcal{T}_{a^k}|$ for different values of $\gamma$ is shown in \ref{tab:listSize}. However, some proposals which have an high IOU overlap with current top-scoring proposal may not be suppressed, however these proposals get suppressed during final second stage of GreedyNMS leading to a huge speed-up with negligible difference in accuracy.

\subsection{Datasets}
We report the detection metrics and latency of our method on COCO \cite{Lin2014MicrosoftCC} and Pascal VOC \cite{everingham2010pascal} datasets.

\textbf{PASCAL VOC: } PASCAL VOC 2007 and 2012 datasets \cite{everingham2010pascal} consist of 20 classes. We report standard mAP at IoU of 0.5 for this dataset. We train our detector on VOC 2007 training set and VOC 2012 training plus validation sets and perform evaluation on the VOC 2007 test set.

\textbf{COCO: } The COCO dataset comprises of 80 classes. We train the models on COCO 2014 training and validation set (minus 2017 validation set) and evaluated our results on 2017 validation set which consists of 5000 images.

\subsection{Speed-Accuracy Trade-off}
The threshold parameter $\gamma$ in ASAP-NMS controls the number of comparisons performed as it affects the size of the anchor list for each anchor template $T^k$. Table \ref{tab:AnchorList} reports the size of the template set $\mathcal{N}_{T^k}$ for each anchor at different thresholds $\gamma$ when the default anchor placement strategy is used, column (R). As expected, $|\mathcal{N}_{T^k}|$ is small for small anchors and much larger for large anchors. Table \ref{tab:AnchorList} also shows the run time for the NMS operation at different thresholds $\gamma$ in milliseconds. The reported run-time includes the time for sorting the input candidate proposals, $\mathcal{S}$ and then running ASAP-NMS on the best proposals $\mathcal{P}$. We use a parallel implementation of sort on the CPU. Finally, Table \ref{tab:AnchorList} reports the mAP on both COCO and VOC datasets for 2 scale inference for different values of $\gamma$.

\begin{table}[t]
\centering
\setlength\tabcolsep{12pt}
\begin{tabular}{c|c|c|c|c|c|}

 \cline{2-6}
 & \multicolumn{3}{c|}{\textbf{COCO}} & \multicolumn{2}{|c|}{\textbf{VOC}}\\
 \cline{2-6}
 & R@0.5 & AP & AP$^{50}$ & AP$^{50}$ & R@0.5 \\
\midrule
\multicolumn{1}{|c|}{Maxpool \cite{Cai2019MaxpoolNMSGR}} & 81.9 & 43.5 & 63.4 & 85.8 & 98.2 \\
\midrule
\multicolumn{1}{|c|}{ASAP} & 89.1 & \textbf{44.2} & \textbf{64.7} & \textbf{86.0} & \textbf{99.3} \\
\bottomrule
\multicolumn{1}{|c|}{Greedy} & \textbf{89.6} & \textbf{44.3} & \textbf{64.8} & \textbf{86.0} & \textbf{99.3} \\
\bottomrule
\end{tabular}
\caption{Comparison of RPN proposal recall and precision at 0.5 IoU. ASAP-NMS uses 300 post-NMS proposals and $\gamma >=0.3$. For Maxpool NMS, we implemented the Multi-Scale variant which has the highest accuracy.}
\label{tab:NMScomparison}
\end{table}

\subsection{Effect of ASAP-NMS on mAP/Recall}
\begin{figure}[!t]
  \centering
  \begin{subfigure}[t]{0.22\textwidth}
    \includegraphics[width=\textwidth]{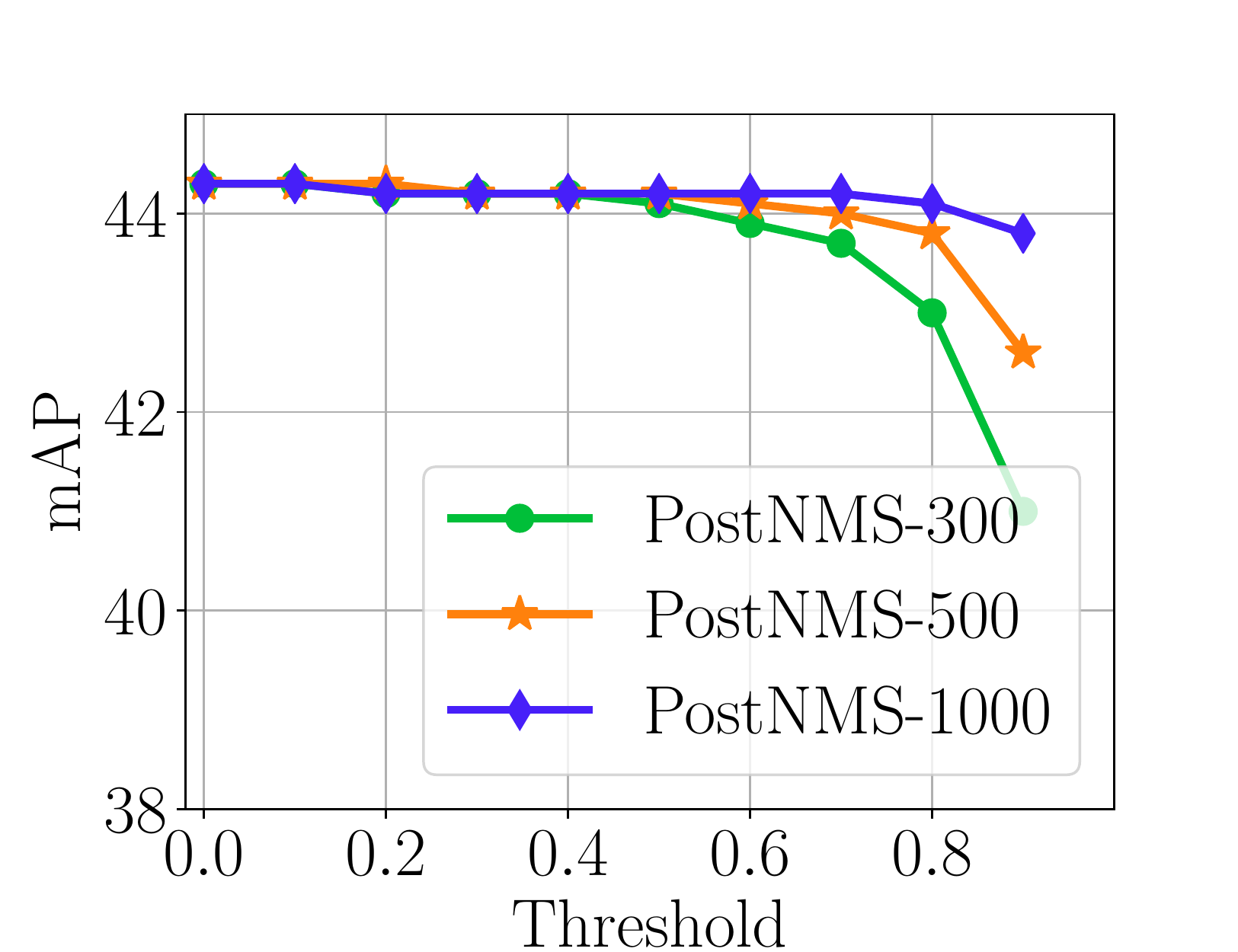}
    \caption{COCO mAP (0.5:0.95)}
    \label{fig:COCO_map_95}
    \end{subfigure}
    \hspace{0.5em}
    \begin{subfigure}[t]{0.22\textwidth}
    \includegraphics[width=\textwidth]{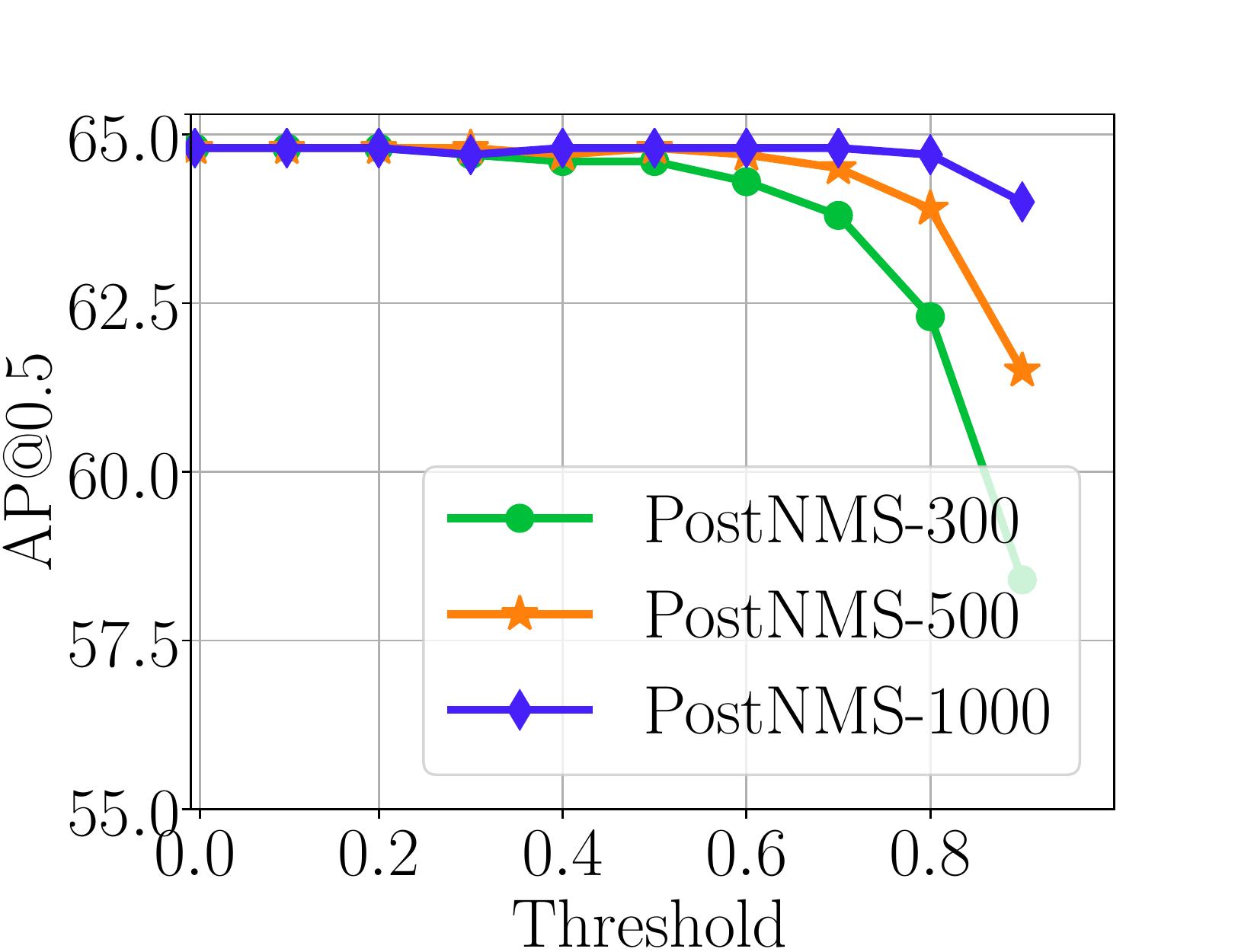}
    \caption{COCO AP@0.5}
    \label{fig:COCO_map_5}
    \end{subfigure}
    \hspace{0.5em}
    \begin{subfigure}[t]{0.22\textwidth}
    \includegraphics[width=\textwidth]{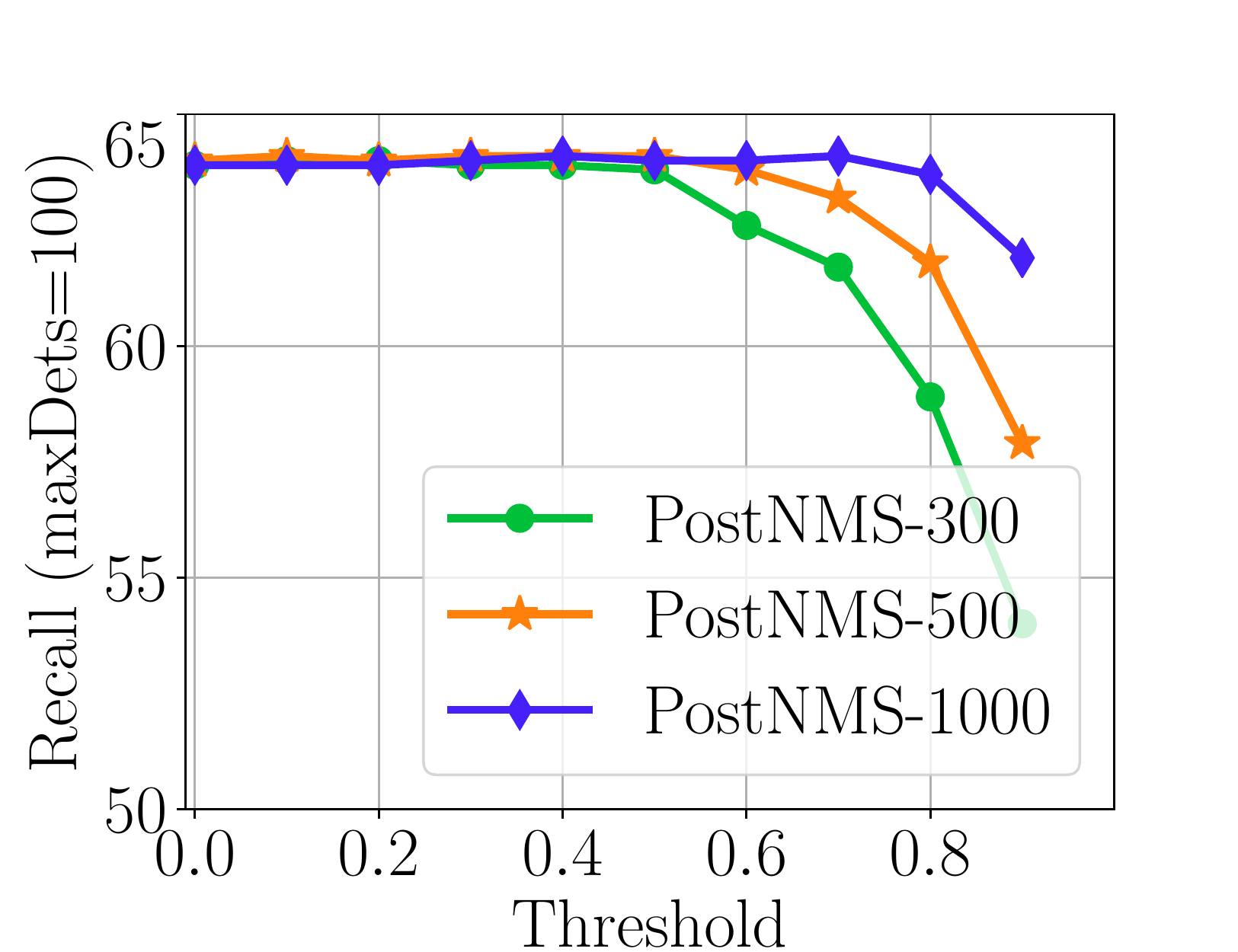}
    \caption{COCO Proposal Recall}
    \label{fig:COCO_recall}
    \end{subfigure}
    \hspace{0.5em}
    \begin{subfigure}[t]{0.22\textwidth}
    \includegraphics[width=\textwidth]{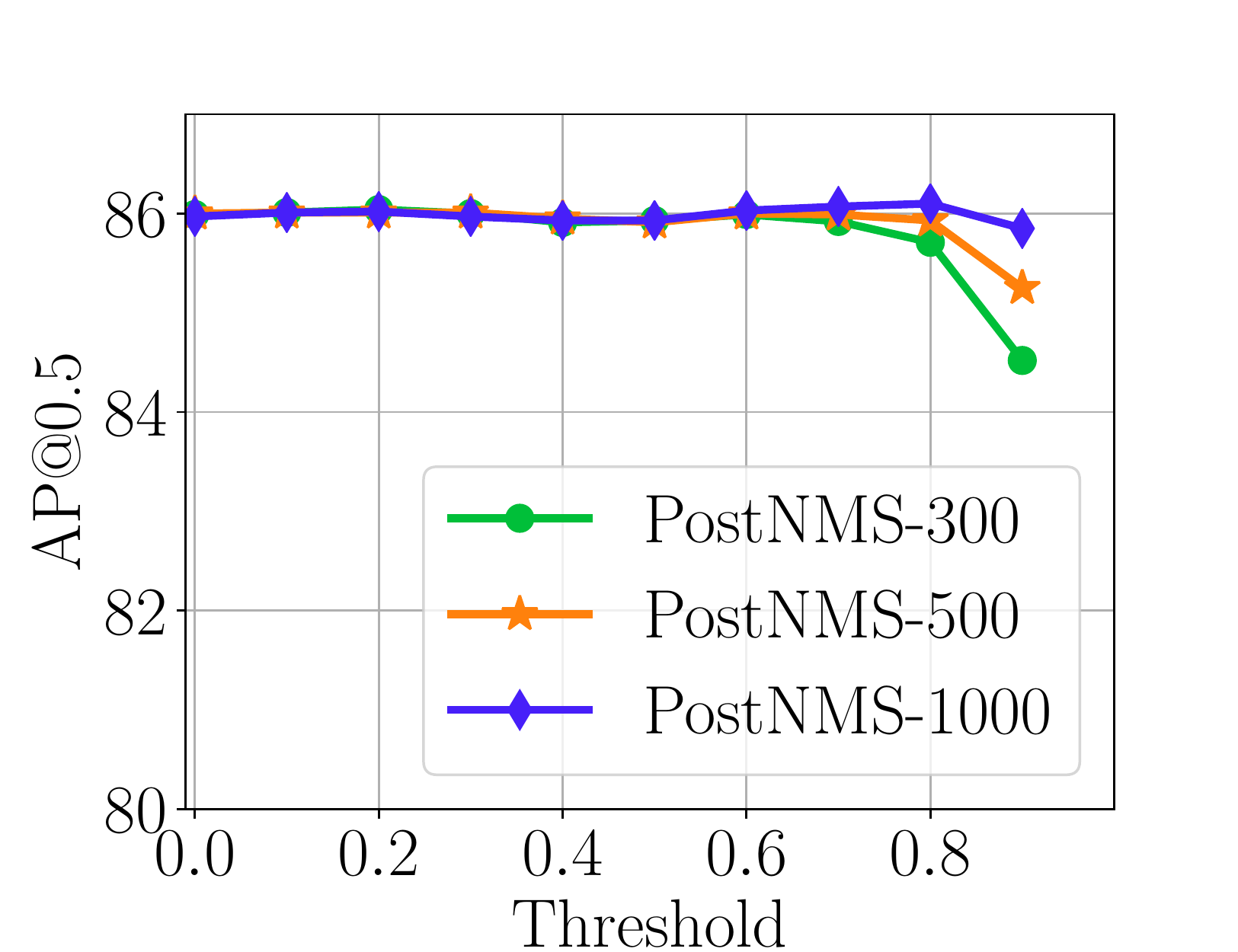}
    \caption{Pascal VOC AP@0.5}
    \label{fig:voc_map}
    \end{subfigure}
\caption{ASAP-NMS with different thresholds and number of post-NMS proposals.}
\label{fig:map_threshold_asap}
\end{figure}

We report the effect of deploying ASAP-NMS with different initial thresholds $\gamma$ on the detection performance. Figures \ref{fig:COCO_map_95}, \ref{fig:COCO_map_5} and \ref{fig:COCO_recall} show the mAP (0.5:0.95), AP@0.5, and the recall on the COCO dataset respectively for different number of post-NMS proposals. As can be seen, even when the total number of post-NMS proposals is limited to 300 and $\gamma$ is varied up to 0.5, we do not see to a noticeable change in recall and mAP. Predictably, the performance starts to drop for larger threshold values. When a higher number of post-NMS proposals is allowed, $\gamma$ as high as 0.8 is also applicable to obtain the same performance as the baseline. The same trend is seen for the PASCAL VOC dataset, as presented in Figure \ref{fig:voc_map}. However, given that this dataset is less challenging, ASAP-NMS  performance does not drop for thresholds as high as 0.8 even when 300 proposals are used. 

\subsection{Comparison with Other NMS Variants}
We compare our algorithm with Greedy-NMS and the recently proposed MaxPool-NMS \cite{Cai2019MaxpoolNMSGR} in Table \ref{tab:NMScomparison}. ASAP-NMS almost matches the recall and mAP performance of Greedy-NMS while being almost 10x faster. It is also better in performance compared to MaxPool-NMS. ASAP-NMS outperforms MaxPool-NMS by more than 0.7\% in terms of COCO AP (0.5:0.95) and 1.2\% for AP at 0.5 overlap. It's proposal recall is also 7.2\% better than Maxpool-NMS on COCO and 1.1\% on PASCAL-VOC. For applications which require high quality proposal recall with a limited candidate set (for dataset annotation, or human in the loop pipelines for high-precision detection), this holds a significant advantage.

\subsection{Strided Anchor Placement}
To reduce $|\mathcal{N}_{T^k}|$, we employ a strided anchor placement strategy and report mAP scores and $|\mathcal{N}_{T^k}|$ in column $S$ in Table \ref{tab:AnchorList}. In this example, we change the stride only for the largest anchor scale ($384\times384$) to 32, which leads to a decrease in $|\mathcal{N}_{T^k}|$ for anchors at the largest two scales. This does not lead to any drop in performance while providing a speedup of 57\% at $\gamma =$ 0.3 and a speedup of 87\% at $\gamma =$ 0.2. Note that at these low levels of $\gamma$, we are almost guaranteed ($< 4\%$) that no NMS errors would happen, as was shown in Table \ref{tab:flipping}. Thus, when using the strided anchor placement scheme ASAP-NMS can default to the exact characteristics of Greedy-NMS while still providing a significant speedup.

\subsection{Qualitative Results}
\begin{figure}[!t]
\centering
\includegraphics[width=0.98\linewidth]{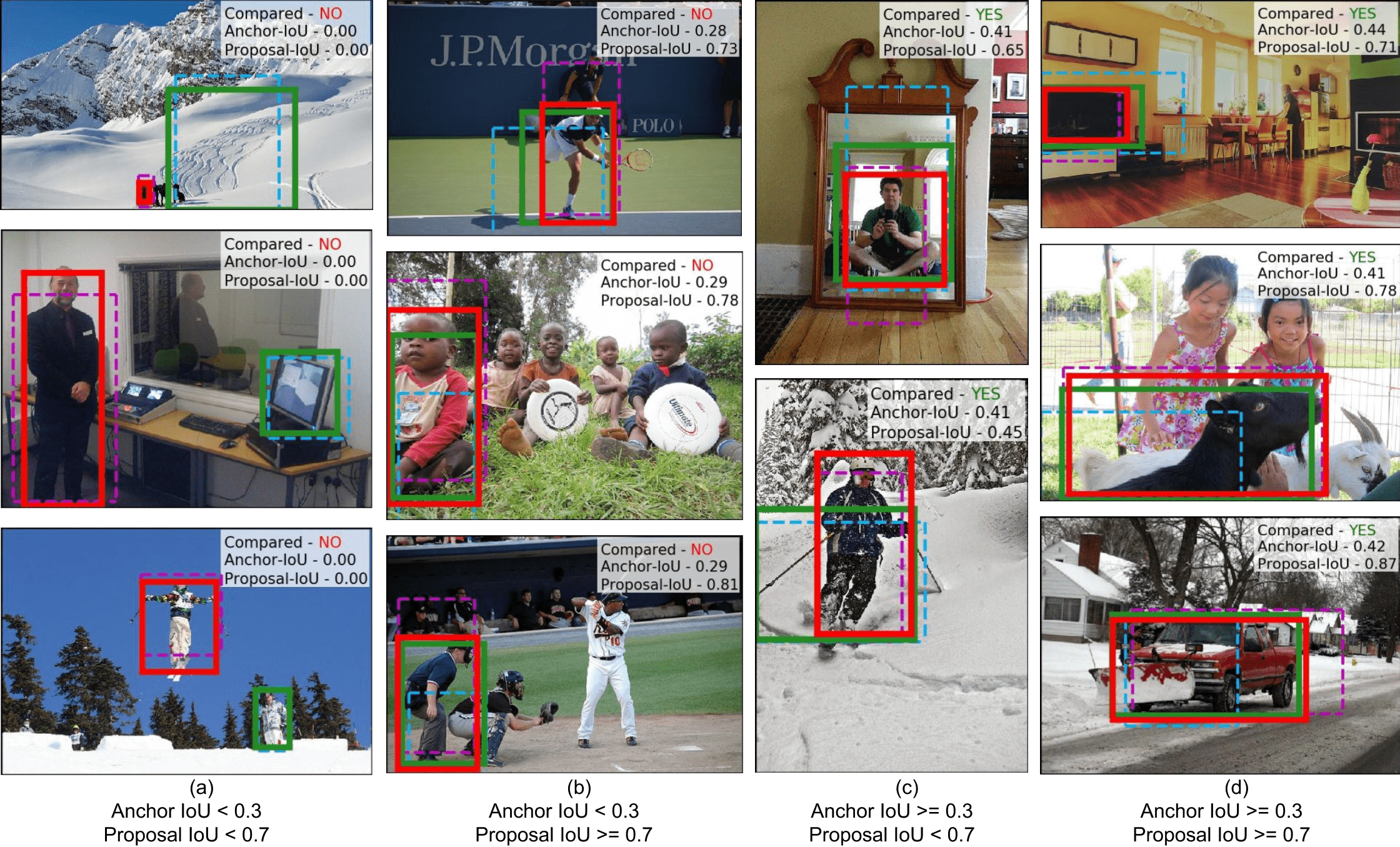}
\caption{Qualitative results from the COCO val2017 data. The high scoring proposal
in the first iteration of ASAP-NMS
is shown in red (it's anchor in purple) and one lower-scoring box is shown in green (it's anchor in blue). ASAP-NMS does not compare the proposals for columns (a) and (b). The probability of situations shown in column (b) is very low ($< 4\%$) with $\gamma=0.3$.}
\label{fig:qualitative_results}
\end{figure}

The objective of ASAP-NMS is to reduce the total number of IoU comparisons while allowing the high scoring proposal to suppress other nearby proposals. In order to visualize the effect of ASAP-NMS, Figure~\ref{fig:qualitative_results} shows pairs of proposals along with ASAP-NMS's decision to compare them or not. The higher and the lower scoring proposals are shown in red and green, respectively. Columns (a) and (b) show examples of pairs for which the comparison is skipped by ASAP-NMS to decrease the overall latency. Column (b) highlights cases where ASAP-NMS failed to suppress the lower-scoring proposal which would otherwise be suppressed by the GreedyNMS. 
%Although visually they look bad, 
As shown in Table \ref{tab:flipping}, for $\gamma=0.3$, the probability of such cases is very low ($< 4\%$). Moreover, the refinement stage NMS, which runs after the final bounding-box regression step will suppress them. Columns (c) and (d) show examples where pairs of proposals are compared by ASAP-NMS.

% How the same idea can improve normal NMS/soft-NMS?
\subsection{Other optimizations for NMS}
The second stage NMS algorithms such as SoftNMS\cite{bodla2017soft} and GreedyNMS need to perform NMS for all $N$ detected boxes and $C$ classes. For a single class, both these algorithms are $\mathcal{O}(N^2)$ as they compute the IoU value for each pair of boxes to check for suppression. Applying NMS per class makes the algorithm, $\mathcal{O}(C*N^2)$, which would become a bottleneck as the number of classes start to increase to a 1000 classes. In practice, this affects SoftNMS-like algorithms more than GreedyNMS because GreedyNMS suppresses many of the boxes in the first few iterations while most implementations of SoftNMS don't suppress any boxes and use the final scores to generate an output.

Inspired by ASAP-NMS, second stage NMS algorithms can be optimized by generating a template per detection which stores the list of detections with significant overlap with it. The order of computation for generating such a mapping is $\mathcal{O}(N^2)$. During the NMS algorithm, for each detection, we only change the score for other detections in the computed template list. If the mean size of the template list for all detections is $M$, the number of times ASAP-NMS checks for suppression becomes $\mathcal{O}(C*N*M)$. The complexity of this algorithm becomes $\mathcal{O}(N^2 + C*N*M)$. As $M$ is typically smaller than $N$, this speeds up second stage NMS with no drop in accuracy.

\begin{table}[t]
\centering
\begin{tabular}{c|c|c|c|c|c|c|c}
% \begin{tabular}{@{}c@{}}Cascade \\ RCNNN\end{tabular}
 &  \begin{tabular}{@{}c@{}} Backbone \\ (fp32 / int8)\end{tabular} & RPN & NMS & ASAP-NMS & \begin{tabular}{@{}c@{}} Head \\ (fp32 / int8)\end{tabular} & Post-process & mAP \\
\bottomrule
ResNet50 & 18.8 / 5.4 & 0.04 & \textbf{5.9} & \textbf{1.2} & 3.6 / 1.1 & 0.5 & 39.0 \\ 
\bottomrule
ResNet101 & 36.8 / 10.5 & 0.03 & \textbf{5.8} & \textbf{1.2} & 3.8 / 1.1 & 0.4 & 40.6
\end{tabular}
\caption{mAP/speed using different backbones for Faster R-CNN on COCO for images of size $800\times1280$ using 300 proposals. Timings (ms) are computed on a single V100 GPU using the popular detectron2 library. The default NMS in detectron2 runs on the GPU whereas ASAP-NMS runs on the CPU.}
\label{tab:detectron2_approx}
\end{table}

\subsection{Latency Compared to Existing Detectors}
In Table \ref{tab:map_speed}, we compare ASAP-NMS with state-of-the-art detectors. SNIPER + ASAP-NMS runs at $25.3$ FPS and obtains an mAP of 44.2\% while SNIPER + Greedy-NMS runs at only $17.5$ FPS on a V100 GPU. Note that we are only measuring the latency, the throughput can be improved further as the GPU is under-utilized with a batch size of 1. Even single-shot detectors like RetinaNet are much slower and achieve a lower mAP. Results for detectors other than SNIPER are taken from the MMDetection repository which reported performance on the same GPU ~\cite{Chen2019MMDetectionOM}. In Table \ref{tab:detectron2_approx}, we share the timings of different components of the Faster-RCNN pipeline with different backbones on V100 GPUs using the popular Detectron2 library by facebook. The numbers clearly demonstrate that even the GPU-optimized NMS used in Detectron2 is a bottle-neck which can be made faster using ASAP-NMS operated at $\gamma=0.4$ even on a CPU. While changing bit precision can make the CNN backbone faster, it does not benefit the greedy-NMS algorithm as it cannot leverage the tensor cores which are optimized for convolutions.

\begin{table}[!t]
\centering
\setlength\tabcolsep{6pt}\begin{tabular}{|c|c|c|c||c|c|}
\cline{2-6}
\multicolumn{1}{c|}{}&  RetinaNet & \begin{tabular}{@{}c@{}}Cascade \\ RCNNN\end{tabular} & \begin{tabular}{@{}c@{}}Cascade \\ Mask RCNNN\end{tabular} &  \begin{tabular}{@{}c@{}}SNIPER \\ +GreedyNMS\end{tabular} & \begin{tabular}{@{}c@{}}SNIPER \\ +ASAP-NMS\end{tabular} \\
\midrule
 mAP &  38.1 & 42.5 & 43.3 & \textbf{44.3} & \textbf{44.2}\\
 \hline
 FPS &   10.9 & 10.3 & 6.8 & 17.5 & \textbf{25.3}\\
 \bottomrule
\end{tabular}
\caption{mAP/speed comparison. mAP is reported for COCO-val2017 and timings are performed on a single V100 GPU using a ResNet-101 backbone.}   
\label{tab:map_speed}
\end{table}
\vspace{-1em}

\section{Conclusion}
Non-Maximum Suppression (NMS) has a crucial role in object detection pipelines to remove the redundancy in the proposal generation stage. However, the widely adapted sequential variant of NMS is a latency bottleneck in the state-of-the-art two-stage detectors. In this paper, we proposed ASAP-NMS, an NMS algorithm which accelerates these detectors by leveraging spatial priors derived from the anchor space. ASAP-NMS noticeably improves the latency of the NMS step from 13.6ms to 1.2ms on a CPU while maintaining the accuracy of a state-of-the-art two-stage object detector. Using ASAP-NMS as a drop-in module during inference, we obtained an mAP of 44.2\% on the COCO dataset while operating at 25FPS on a V100 GPU.

%This article proposed to exploit the spatial priors in the form of anchor's neighbourhood set to reduce the comparisons between distant proposals during NMS. \vsingla{we need to re-word this. we shouldn't be writing the latency drop, as it is for single scale, and mAP is for 2 scales, at least it needs to be mentioned properly. I'm commenting this section till we fix this.} 
% We demonstrated an efficient lookup table based implementation scheme to obtain the neighourhood set from a pre-computed table that reduced the latency of NMS from 13.6ms to 1.7ms without any drop in accuracy on MS-COCO dataset.
% We report an mAP of 44.2\%@20Hz using a state-of-the-art two-stage object-detector on MS-COCO.

\bibliographystyle{splncs04}
\bibliography{egbib}
\end{document}